\pdfoutput=1
\documentclass[11pt]{article}

\usepackage[final]{acl}

\usepackage{times}
\usepackage{latexsym}
\usepackage{adjustbox}
\usepackage[T1]{fontenc}
\usepackage[utf8]{inputenc}

\usepackage{microtype}

\usepackage{inconsolata}

\usepackage{graphicx}

\usepackage{multirow}
\usepackage{colortbl}
\usepackage{array}
\usepackage{xurl}
\newcolumntype{L}[1]{>{\raggedright\let\newline\\\arraybackslash\hspace{0pt}}m{#1}}
\newcolumntype{C}[1]{>{\centering\let\newline\\\arraybackslash\hspace{0pt}}m{#1}}
\newcolumntype{R}[1]{>{\raggedleft\let\newline\\\arraybackslash\hspace{0pt}}m{#1}}

  {\list{}{\leftmargin=0.0in\rightmargin=0.0in}  \item[]  }%
  {\endlist}

\title{Just Go Parallel: Improving the Multilingual Capabilities of \\ Large Language Models}

\author{Muhammad Reza Qorib, Junyi Li, \and Hwee Tou Ng \\
        Department of Computer Science, National University of Singapore \\
        \texttt{mrqorib@u.nus.edu, junyi\_cs@nus.edu.sg, nght@comp.nus.edu.sg}}

\begin{document}
\maketitle
\begin{abstract}
Large language models (LLMs) have demonstrated impressive translation capabilities even without being explicitly trained on parallel data. This remarkable property has led some to believe that parallel data is no longer necessary for building multilingual language models. While some attribute this to the emergent abilities of LLMs due to scale, recent work suggests that it is actually caused by incidental bilingual signals present in the training data. Various methods have been proposed to maximize the utility of parallel data to enhance the multilingual capabilities of multilingual encoder-based and encoder-decoder language models. However, some decoder-based LLMs opt to ignore parallel data instead. In this work, we conduct a systematic study on the impact of adding parallel data on LLMs' multilingual capabilities, focusing specifically on translation and multilingual common-sense reasoning. Through controlled experiments, we demonstrate that parallel data can significantly improve LLMs' multilingual capabilities.\footnote{Source code, checkpoints, and data are available at: \url{https://github.com/nusnlp/just-go-parallel}}
\end{abstract}

\section{Introduction}
To democratize the benefits of large language models (LLMs) for the whole world, many initiatives have been undertaken to build LLMs that possess multilingual capabilities \cite{bloom, DBLP:journals/corr/abs-2308-16149}. Multilingual capabilities enhance the accessibility and inclusivity of the model and help reduce its inherent bias \cite{zhu2024multilinguallargelanguagemodels, 10.1145/3597307}.

Even without being explicitly trained with parallel data, LLMs are reported to have impressive translation capabilities \cite{Radford2019LanguageMA, lin-etal-2022-shot}. \citet{briakou-etal-2023-searching} reported that the translation capabilities of LLMs are highly correlated with the bilingual signals in their training data, particularly translation pairs. The presence of bilingual signals is often incidental, meaning they are not deliberately added to the training data. When these bilingual signals are removed, LLMs lose the capability to translate between English and languages with non-Latin scripts.

While some multilingual LLMs are trained with parallel data \cite{alves2024tower}, some are not \cite{bloom}, despite having parallel data sources such as OPUS-100 \cite{zhang-etal-2020-improving} or EuroParl \cite{koehn-2005-europarl} in their training data.
While parallel texts are not always available especially for extremely low-resource languages, the situation today has improved dramatically compared to a decade ago. Due to efforts such as NLLB (No Language Left Behind; \citealt{costa2022no}), many more languages -- even for languages that were previously considered low-resource like Indonesian -- now have enough publicly available parallel texts to make a difference in building multilingual large language models, as we demonstrate in this paper.

Many encoder-based \cite{ConneauL19, ouyang-etal-2021-ernie} and encoder-decoder \cite{liu-etal-2020-multilingual-denoising, chi-etal-2021-mt6} language models aim to improve their multilingual capabilities through parallel corpora. Much work has been proposed to specifically enhance cross-lingual alignments \cite{Cao2020Multilingual, luo-etal-2021-veco}, even when parallel sentences are unavailable \cite{lu-etal-2023-take}. On the other hand, decoder language models often deliberately ignore the parallel data available in their training corpus, such as by not including the English portion of the parallel data or randomly mixing all the training data together. This situation warrants a systematic investigation of the effect of parallel data on large language models' multilingual capabilities.

In this research, we conduct a systematic study to investigate whether adding parallel data helps in enhancing a large language model's multilingual capabilities and what the best strategy is for incorporating them into a language model's training data.

Our main contributions are as follows.
\begin{enumerate}
\item We report that adding parallel data to the training data is more effective than adding unrelated monolingual corpora of other languages for enhancing the multilingual capabilities of decoder-based large language models.
\item We find that training the model with parallel data at the end of the training process is the most effective approach for improving multilingual performance, while training at the beginning of the process leads to serious catastrophic forgetting and wastes the parallel data.
\item We report that LLMs do not have the ability to perform translation in the opposite direction of what they were trained on.
\item We show that the amount of bilingual signal in the training data affects LLMs' translation capability.
\item We release the code, checkpoints, and training data of our models to facilitate further study.
\end{enumerate}

\section{Related Work}

\subsection{Multilingual Language Models}

Benefiting from the complete open-sourcing of data and code from many LLMs, such as BLOOM~\cite{bloom}, LLM360~\cite{llm360}, and Falcon~\cite{falcon}, many studies have begun developing multilingual language models based on these LLMs by incorporating multilingual data for continual pre-training~\cite{ConneauL19,XueCRKASBR21,lin-etal-2022-shot}. Although recent instruction-following LLMs are primarily pre-trained and fine-tuned on a limited number of resource-rich languages, they have demonstrated substantial multilingual comprehension and generation capabilities~\cite{llama2,WangCZLSL24,NiklausMRGSC23}. However, due to the limitation of imbalanced training data distribution~\cite{abs-2305-18098}, these multilingual language models still fall short in those languages with scarce resources~\cite{abs-2404-04925}. To understand how language models acquire multilingual capabilities, \citet{abs-2404-19159} investigate factors influencing the performance of multilingual language models and reveal that script type and language family are critical for unseen languages, highlighting the importance of cross-lingual transfer learning. Additionally, research by \citet{TangLH0WZWW24} aims to explain the underlying mechanisms by which LLMs process multilingual texts. Their research indicates that the proficiency of LLMs in processing languages is predominantly due to a small subset of neurons, primarily situated in the models' top and bottom layers.

\subsection{Enhancing Multilingual Capabilities} 

To enhance the multilingual capabilities of LLMs, one line of work is cross-lingual transfer, where the capability of one language can be transferred to other languages~\cite{HuangTZZSXW23,EtxanizASLA24,abs-2311-08097}. By designing cross-lingual alignment prompting that instructs LLMs to self-translate a question into another language~\cite{0001CWHC23} or utilize an external machine translation system~\cite{abs-2401-01055}, the capabilities of language generation and instruction-following can be transferred to a non-English language. Besides, several efforts have been devoted to knowledge distillation on synthetic data from high-resource languages to low-resource ones~\cite{abs-2401-07037,abs-2407-19610,ZhangWLWWL0024}. Another line of work is cross-lingual alignment, which involves constructing alignment data and loss functions to align mid- and low-resource
languages with those that are resource-rich~\cite{SchusterRBG19,Wen-YiM23,ZhuHYSCB24}. For example, \citet{Chi0ZHMHW20} introduce a denoising word alignment pre-training task that predicts masked tokens in another language. \citet{abs-2401-05811} leverage the capabilities of LLMs to translate previously unsupported languages for building aligned data, overcoming the weak cross-lingual signals caused by data scarcity.

\section{Method}
Large language models are often trained on internet-sourced corpora predominantly in English~\cite{biderman2023pythia, groeneveld-etal-2024-olmo}. We investigate whether incorporating parallel data enhances multilingual capabilities by comparing it to training without parallel data or with monolingual corpora from other languages. Our objective is to empirically examine how the inclusion of parallel data in the training set affects multilingual LLMs' performance across multiple languages.  

To ensure a controlled comparison, we maintain the order and quantity of training data across all experiments. When parallel data are added to the training set, an equivalent amount of non-parallel data from the last portion is removed to preserve a consistent training set size and order.

We also explore the optimal placement of parallel data within the language model's training process. Specifically, we experiment with inserting parallel data at the beginning of training, distributing it throughout the training data, and adding it at the end. We define seven experimental settings: \textsc{No Parallel}, \textsc{Multilingual},  \textsc{Parallel Non-Adjacent}, \textsc{Parallel First}, \textsc{Parallel Distributed}, \textsc{Parallel Last (all)}, and \textsc{Parallel Last (uni)}, as detailed below.

\subsection{No Parallel}  
In this setting, no parallel data are added to the training data. This approach mirrors the typical strategy used for constructing English-centric LLMs, such as Pythia~\cite{biderman2023pythia} and TinyLlama~\cite{zhang2024tinyllama}. Although no parallel data are intentionally included, the training data may incidentally contain some parallel texts sourced from the internet. However, according to our language detection analysis (Table~\ref{tab:lang_percentage}), the amount of such data is minimal.

\subsection{Multilingual}  
In this setting, monolingual data from other languages are distributed uniformly throughout the non-parallel data. This approach resembles common strategies for building multilingual language models by incorporating monolingual corpora from multiple languages~\cite{lu-etal-2023-take, bloom}. To control for the choice of text, the monolingual data added are derived from the non-English half of the parallel data. That is, in our experiments, monolingual data in Indonesian and Chinese are added in this setting, and the equivalent amount of English data from the last portion of the \textsc{No Parallel} setting is removed. Note that the English half of the parallel data are not included in this setting. This setup evaluates whether a language model can learn cross-lingual mappings independently without explicitly aligning semantically equivalent sentences across languages.

\subsection{Parallel Non-Adjacent}  
In this setting, parallel data for all language pairs are uniformly distributed throughout the non-parallel data. However, an English sentence and its translation are not placed next to each other. Instead, the English sentences in the parallel data are randomly shuffled, such that each non-English sentence from the parallel data is followed by a random English sentence. This setup evaluates whether the presence of semantically equivalent sentences in the training data, but without explicitly placing a sentence next to its translation, helps in learning cross-lingual mappings.

\subsection{Parallel First}  
In this setting, parallel data for all translation directions are introduced at the beginning of training. The parallel data are presented as adjacent sentence pairs, where a sentence and its translation are placed next to each other. The rationale behind this setup is that early exposure to parallel data may help the model establish cross-lingual mappings, enabling it to better leverage incidental bilingual signals present in the non-parallel training data.  

\subsection{Parallel Distributed}  
In this setting, parallel data for all translation directions, presented as adjacent sentence pairs, are distributed throughout the non-parallel data. Since large language models are believed to acquire multilingual capabilities from bilingual signals~\cite{briakou-etal-2023-searching}, particularly translation pairs, this setup aims to amplify such signals.  

\subsection{Parallel Last (all)}  
In this setting, parallel data for all translation directions, presented as adjacent sentence pairs, are added at the end of training. This approach can also be interpreted as second-stage training, where the model receives bilingual exposure after being pre-trained primarily on English data.

\subsection{Parallel Last (uni)}  
This setting is similar to \textsc{Parallel Last (all)}, except that the model is trained on only one translation direction (e.g., English to Chinese). For each translation direction, a separate model is trained, resulting in specialized models rather than a general multilingual LLM. 

\begin{table}
  \centering
  \begin{tabular}{lrrr}
    \hline
    \textbf{Corpus} &  \textbf{\# sents} &  \multicolumn{2}{c}{\textbf{\# tokens}} \\
    {} & {} & \textbf{Chinese} & \textbf{English} \\
    \hline
    ParaCrawl &  14.2M & 620.6M & 357.9M \\
    NewsComm & 0.3M & 19.8M & 10.3M \\
    Wiki Titles & 0.9M & 6.1M & 10.9M \\
    UN Parallel & 15.9M & 999.4M & 579.9M \\
    WikiMatrix & 2.6M & 150.0M & 83.5M \\
    \hline
    Total & 33.9M & 1,795.9M & 1,042.5M \\
    \hline
  \end{tabular}
  \caption{\label{zh-parallel}
    Statistics of Chinese-English training parallel data.
  }
\end{table}

\begin{table}
  \centering
  \begin{tabular}{lrrr}
    \hline
    \textbf{Corpus} &  \textbf{\# sents} &  \multicolumn{2}{c}{\textbf{\# tokens}} \\
    {} & {} & \textbf{Indonesian} & \textbf{English} \\
    \hline
    Total & 54.1M & 1,222.0M & 883.5M \\
    \hline
  \end{tabular}
  \caption{\label{id-parallel}
    Statistics of Indonesian-English training parallel data.
  }
\end{table}

\section{Experiments}
\label{sec:exp}
We build our model based on TinyLlama, a 22-layer transformer decoder LLM with 1.1B parameters. We use a subset of SlimPajama~\cite{cerebras2023slimpajama} as our non-parallel training data. When referring to non-parallel data, we mean SlimPajama data, which are predominantly in English and do not deliberately include parallel data. The subset of SlimPajama used consists of 82.35\% English, 0.19\% Indonesian, 0.12\% Chinese, and 17.34\% other languages (See Table~\ref{tab:lang_percentage} in the Appendix).  

We measure the language model's multilingual capabilities by evaluating its translation and common-sense reasoning performance. In our experiments, we focus on English (EN), Simplified Chinese (ZH), and Indonesian (ID) as case studies. We selected Chinese for its script diversity and Indonesian for its mid-resource status. To incorporate parallel data into our training set, we use widely adopted parallel corpora for machine translation tasks. Specifically, we use the training data from the WMT-2022 general (news) machine translation task~\cite{kocmi-etal-2022-findings} for Chinese-English translation pairs, excluding the CCMT corpus\footnote{The CCMT corpus requires registration, but we did not receive a response after submitting our registration request.} (Table~\ref{zh-parallel}). For Indonesian-English translation pairs, we use the training data from the WMT-2021 large-scale multilingual machine translation task~\cite{wenzek-etal-2021-findings} (Table~\ref{id-parallel}).  

For each translation pair, we format the parallel data as plain text using the template:  
``\texttt{\{source language\}: \{source sentence\}\textbackslash n\{target language\}: \{target sentence\}}''.  
This approach is inspired by the format of incidental parallel data found in PALM's~\cite{10.5555/3648699.3648939} training set, as reported by~\citet{briakou-etal-2023-searching}. By default, we alternate the translation direction of the language pairs (e.g., EN $\rightarrow$ ID, ID $\rightarrow$ EN, ZH $\rightarrow$ EN, EN $\rightarrow$ ZH). Text sequences are concatenated using the end-of-sentence pseudo-token \texttt{<\textbackslash s>} and split into chunks, each with a size equal to eight times the context window. Leftover text segments that are shorter than this are discarded. After pre-processing, the parallel data amount to 4.5B tokens\footnote{Throughout this paper, the number of tokens is consistently measured using the TinyLlama tokenizer, which shares the same vocabulary as the Llama 2 models.}.

We train the model on up to 167B tokens\footnote{All experimental settings are trained on 167B tokens, except the \textsc{Parallel Last (all)} (166B tokens) and \textsc{Parallel Last (uni)} (164B tokens) settings. The \textsc{Parallel Last (all)} setting uses 162B tokens of non-parallel data and 4B tokens of parallel data (due to saving the checkpoints every 1B tokens, not all parallel data are consumed), while the \textsc{Parallel Last (uni)} setting uses 162B tokens of non-parallel data and 2B tokens of parallel data (EN$\leftrightarrow$ZH or EN$\leftrightarrow$ID parallel data only, instead of all available parallel data).} using NVIDIA H100 GPUs. Due to the high computational cost, each experiment is conducted only once. We save checkpoints every 5,000 steps (\textasciitilde 5.2B tokens), but for the first 5,000 steps of the \textsc{Parallel First} setting and the last 5,000 steps of the \textsc{Parallel Last} settings, we save checkpoints every 1,000 steps to analyze the effects of parallel data in a more fine-grained manner.  

After training, we select the checkpoint with the highest average BLEU score~\cite{papineni-etal-2002-bleu} across all translation directions on the development set. We use the WMT-2022 test set for Chinese-to-English and English-to-Chinese translation and the Flores-200 dev set~\cite{costa2022no} for Indonesian-to-English and English-to-Indonesian translation as the development set.  

\begin{table*}[thb]
\centering
\begin{tabular}{lrrrrr}
\hline
\hline
\textbf{Model} & \textbf{Param} & \textbf{EN $\rightarrow$ ID} & \textbf{ID $\rightarrow$ EN} & \textbf{EN $\rightarrow$ ZH} & \textbf{ZH $\rightarrow$ EN} \\
\hline
\hline
\multicolumn{6}{c}{Zero-Shot}                              \\
\hline
No Parallel    & 1.1B & 2.49     & 1.52     & 0.80     & 1.30     \\
Multilingual & 1.1B & 2.38 & 5.92 & 0.81 & 3.72 \\ 
Parallel Non-Adjacent   & 1.1B & 1.98 & 14.69 & 1.01 & 4.50     \\
Parallel First & 1.1B & 7.42     & 5.57     & 9.64     & 2.71     \\
Parallel Distributed   & 1.1B & 21.95    & 27.48    & 12.08    & 7.40     \\
Parallel Last (all)  & 1.1B & 35.91    & 35.36    & 9.62     & 10.73    \\
Parallel Last (uni)  & 1.1B & 44.19    & 41.91    & 28.51     & 16.10    \\
\hline
BLOOM          & 1.1B & 2.19      & 18.39      & 2.27      & 4.58      \\
NLLB          & 1.3B & 44.64      & 43.06      & 27.58      & 19.25      \\
\hline
\hline
\multicolumn{6}{c}{Few-Shot (5 examples)}                                 \\
\hline
No Parallel    & 1.1B & 2.60     & 0.97     & 0.79     & 0.83      \\
Multilingual   & 1.1B & 2.49     & 9.75     & 1.73     & 4.25     \\
Parallel Non-Adjacent   & 1.1B & 3.31 & 13.90 & 3.76 & 2.69     \\
Parallel First & 1.1B & 25.41    & 21.02    & 18.61    & 7.09     \\
Parallel Distributed   & 1.1B & 31.80    & 33.66    & 23.21    & 12.51    \\
Parallel Last (all)  & 1.1B & 41.51    & 39.08    & 32.61    & 15.20    \\
Parallel Last (uni)  & 1.1B & 44.32    & 41.60    & 33.31    & 16.87    \\
\hline
BLOOM          & 1.1B & 21.92     & 27.29      & 18.10      & 10.72      \\
\hline
\hline
\end{tabular}
  \caption{\label{tab:result}
    Translation performance (BLEU scores) of each experimental setting, along with the number of parameters in the model (\textsc{Param}). NLLB is specialized for machine translation, so we do not evaluate its few-shot performance.
  }
\end{table*}

\subsection{Evaluation}  
We assess the model's translation performance using the BLEU score metric from SacreBLEU~\cite{post2018call}. Machine translation performance is evaluated on the Chinese-to-English and English-to-Chinese test sets of WMT-2023\footnote{\url{https://www.statmt.org/wmt23/translation-task.html}}~\cite{semenov-etal-2023-findings} and on the devtest set of Flores-200 for Indonesian-to-English and English-to-Indonesian. Since Flores-200 does not provide test sets for both directions, we reverse the translation directions during evaluation. We have verified that none of the evaluation data are included in the training data, ruling out the possibility of data leakage.

We perform zero-shot and few-shot evaluations using 5 examples. The evaluation follows the code and prompt style of ALMA~\cite{xu2024a} (Table~\ref{tab:zero_prompt_example} in the Appendix). Statistical significance is measured using the paired approximate randomization method with 10,000 trials and a significance threshold ($p$-value) of 0.05.  

We also evaluate the model's common-sense reasoning performance in English, Chinese, and Indonesian using a zero-shot approach across several benchmarks. For English, we test on ARC (Easy and Challenge)~\cite{clark2018think}, HellaSwag~\cite{hellaswag}, BoolQ~\cite{boolq}, OpenBookQA~\cite{openbookqa}, PIQA~\cite{piqa}, SciQ~\cite{sciq}, and WinoGrande~\cite{winogrande}. For Chinese, we use the Chinese subsets of XWinograd~\cite{xwinograd}, XStoryCloze~\cite{lin-etal-2022-shot}, XNLI~\cite{xnli}, and XCOPA~\cite{xcopa}. Since XWinograd and XNLI do not have Indonesian subsets, Indonesian common-sense reasoning is evaluated using the Indonesian subsets of XStoryCloze and XCOPA. We utilize the Language Model Evaluation Harness (LM-Eval) framework\footnote{\url{https://github.com/EleutherAI/lm-evaluation-harness/}} for common-sense reasoning evaluation.  

\begin{table*}[thb]
\centering
\begin{adjustbox}{max width=\textwidth}
\begin{tabular}{lrrrrrrrrr}
\hline
\hline
\textbf{Model} & \textbf{ARC\textsubscript{C}} & \textbf{ARC\textsubscript{E}} & \textbf{BoolQ} & \textbf{HS} & \textbf{OBQA} & \textbf{PIQA} & \textbf{SciQ} & \textbf{WG} & \textbf{Avg} \\
\hline
\hline
No Parallel    & 25.51     & 45.08     & 60.12     & 46.30 & 32.00 & 68.66 & 72.50 & 53.35 & 50.44    \\
Multilingual   & 27.30     & 46.93     & 57.31     & 48.25 & 32.40 & 69.15 & 74.00 & 53.91 & 51.16 \\
Parallel Non-Adjacent   & 25.43 & 46.21 & 60.15 & 47.67 & 30.20 & 68.23 & 73.20 & 54.06 & 50.64 \\
Parallel First & 21.84     & 30.56     & 41.07     & 25.73 &  25.00 & 52.88 & 44.60 & 50.28 & 36.50   \\
Parallel Distributed   & 26.45    & 47.05    & 60.61    & 46.49 & 31.60 & 68.88 & 76.30 & 53.28 & 51.33   \\
Parallel Last (all) & 24.40    & 41.79    & 60.21     & 42.21 & 30.20 & 66.10 & 73.80 & 53.91 & 49.08   \\
\hline
BLOOM          & 25.60    & 45.41      & 59.11    & 42.97 & 29.40 & 67.25 & 74.60 & 55.01 & 49.92 \\
\hline
\hline
\end{tabular}
  \end{adjustbox}
  \caption{\label{tab:en-commonsense-result}
    English common-sense reasoning performance in each experimental setting, measured based on the model's accuracy on ARC Challenge, ARC Easy, BoolQ, HellaSwag, OpenBookQA, PIQA, SciQ, and WinoGrande.
  }
\end{table*}

\begin{figure}[htb]
\centering
\includegraphics[width=0.99\linewidth]
{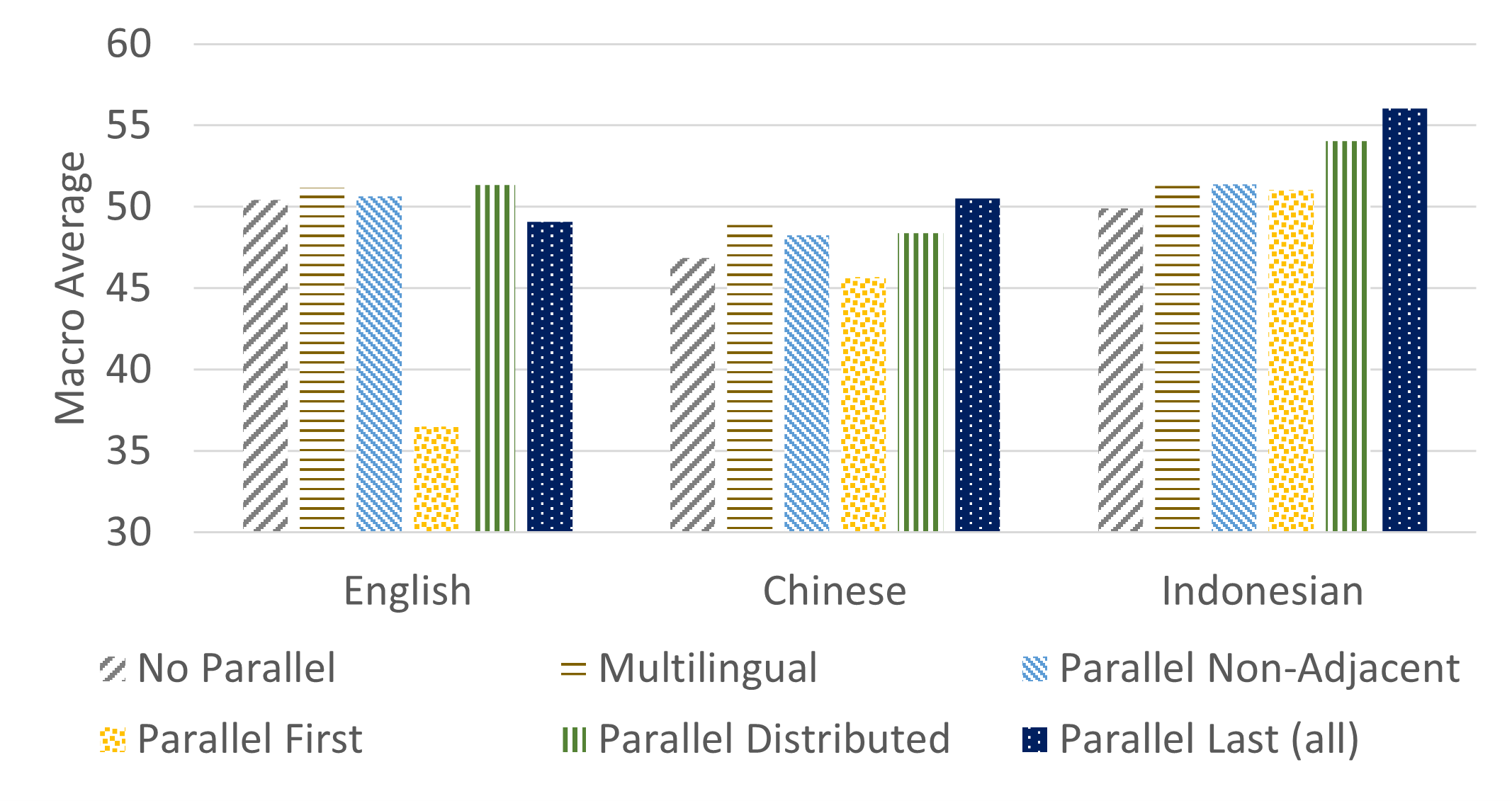}
\caption{Macro-average of the common-sense reasoning benchmarks.}
\label{fig:cse}
% \vspace*{-0.5\baselineskip}
\end{figure}

\section{Results}
We find that training the model with parallel data at the end yields the best translation performance, especially when the model is trained in only one language direction (Table~\ref{tab:result}). Adding parallel data at the beginning provides only a minor improvement over the \textsc{No Parallel} strategy in zero-shot evaluation, but the difference becomes much more pronounced in few-shot evaluation. Few-shot inference also greatly improves the scores of \textsc{Parallel Distributed} and \textsc{Parallel Last (all)}, narrowing the gap between \textsc{Parallel Last (all)} and \textsc{Parallel Last (uni)}.  
Notably, the \textsc{Parallel Distributed} setting significantly outperforms \textsc{Multilingual} and \textsc{Parallel Non-Adjacent}. The \textsc{Parallel Distributed} and \textsc{Parallel Non-Adjacent} settings are trained on the same data for the same duration, differing only in how the parallel data are placed in the training data. The \textsc{Parallel Distributed} setting even outperforms BLOOM, despite BLOOM being trained with more Indonesian and Chinese data -- 20 GB of Indonesian and 261 GB of Chinese texts, compared to less than 5 GB of texts in each language in our setup. This highlights the effectiveness of parallel data in enhancing LLMs' translation capabilities.  

Beyond translation, the \textsc{Parallel Distributed} setting also outperforms the \textsc{Multilingual} and \textsc{Parallel Non-Adjacent} settings on English and Indonesian common-sense reasoning benchmarks while maintaining comparable performance on Chinese common-sense reasoning benchmarks (Figure~\ref{fig:cse}). The \textsc{Parallel Last  (all)} setting significantly outperforms the other experimental settings in Chinese and Indonesian, although it shows a slight decline in English performance, as shown in Table~\ref{tab:en-commonsense-result}. Detailed scores for Chinese and Indonesian common-sense reasoning are provided in Table~\ref{tab:zh-commonsense-result} and Table~\ref{tab:id-commonsense-result}, respectively, in the Appendix. These findings indicate that parallel data not only improves translation performance but also enhances LLMs' common-sense reasoning abilities in non-English languages.

\section{Discussions}
\subsection{All Directions vs. Unidirectional Translation}

As shown in Table~\ref{tab:result}, training with unidirectional data leads to higher translation performance compared to training with data in all translation directions, although the unidirectional approach creates specialized models instead of a single multilingual model. These models cannot translate languages they are not trained on, and surprisingly, they also fail to translate in the opposite direction of the language pair they are trained on (Table~\ref{tab:uni_result}). Their performance is very poor, even worse than models trained without parallel data. We found that they often produce nonsensical outputs due to this limitation. This phenomenon may be related to the issue of autoregressive LLMs struggling with inverse relationships, dubbed the \textit{reversal curse} \cite{reversal_curse}.

Adding few-shot examples does not help unidirectional models to translate other languages or the opposite direction of the same language pair. Even in the same translation direction as their training data, the performance gain is modest. In contrast, the few-shot performance of the \textsc{Parallel Last (all)} model is quite close to the best performance of each unidirectional model while retaining the ability to translate between all language pairs.  

On the common-sense reasoning task, models trained with unidirectional parallel data directed \textit{into} the evaluated language perform slightly worse than the model trained with parallel data in all directions (\textsc{Parallel Last (all)}). For English, models trained only on ID $\rightarrow$ EN or ZH $\rightarrow$ EN parallel data achieve lower scores than the \textsc{Parallel Last (all)} model (Table~\ref{tab:uni_cs}). A similar trend holds for Chinese and Indonesian: training on only EN $\rightarrow$ ZH and EN $\rightarrow$ ID parallel data results in worse performance on Chinese and Indonesian reasoning tasks, respectively.

While unidirectional training may not be suitable for building multilingual large language models with general capabilities, it is highly effective for developing machine translation models. By fine-tuning a large language model (pretrained primarily on English data) with just one epoch of parallel data, we can achieve very high translation performance—exceeding a BLEU score of 41—especially for languages that use the same script as English, such as Indonesian.

\begin{table}[thb]
\setlength\tabcolsep{1.5pt}
\centering
\begin{tabular}{p{0.3\linewidth} | R{0.14\linewidth} R{0.14\linewidth} R{0.14\linewidth} R{0.14\linewidth} }
\hline
{}  & \multicolumn{4}{c}{Translation Evaluation} \\
\cline{2-5}
Parallel Data & \multicolumn{1}{L{0.14\linewidth}}{EN $\rightarrow$ ID} & \multicolumn{1}{L{0.14\linewidth}}{ID $\rightarrow$ EN} & \multicolumn{1}{L{0.14\linewidth}}{EN $\rightarrow$ ZH} & \multicolumn{1}{L{0.14\linewidth}}{ZH $\rightarrow$ EN} \\
\hline
\multicolumn{5}{c}{Zero-Shot} \\
\hline
All directions & \cellcolor[HTML]{82C77D}35.91 & \cellcolor[HTML]{84C87D}35.36 & \cellcolor[HTML]{E1E383}9.62  & \cellcolor[HTML]{DDE283}10.73 \\
EN $\rightarrow$ ID       & \cellcolor[HTML]{64BF7C}44.19 & \cellcolor[HTML]{F8716C}0.07  & \cellcolor[HTML]{FCC17B}0.77  & \cellcolor[HTML]{F9816F}0.21  \\
ID $\rightarrow$ EN       & \cellcolor[HTML]{F86B6B}0.02  & \cellcolor[HTML]{6CC17C}41.91 & \cellcolor[HTML]{F98570}0.25  & \cellcolor[HTML]{F86C6B}0.03  \\
EN $\rightarrow$ ZH       & \cellcolor[HTML]{F8736C}0.09  & \cellcolor[HTML]{FBAC77}0.59  & \cellcolor[HTML]{9DCF7F}28.51 & \cellcolor[HTML]{F86A6B}0.01  \\
ZH $\rightarrow$ EN       & \cellcolor[HTML]{F8696B}0.00  & \cellcolor[HTML]{FAEA84}2.73  & \cellcolor[HTML]{F8776D}0.13  & \cellcolor[HTML]{C9DC81}16.10 \\
None           & \cellcolor[HTML]{FBEA84}2.49  & \cellcolor[HTML]{FEEB84}1.52  & \cellcolor[HTML]{FCC47C}0.80  & \cellcolor[HTML]{FFEB84}1.30  \\
\hline
\multicolumn{5}{c}{Few-Shot (5 examples)} \\
\hline
All directions & \cellcolor[HTML]{6EC17C}41.51 & \cellcolor[HTML]{76C47D}39.08 & \cellcolor[HTML]{8ECB7E}32.61 & \cellcolor[HTML]{CDDD82}15.20  \\
EN $\rightarrow$ ID       & \cellcolor[HTML]{63BE7B}44.32 & \cellcolor[HTML]{F8796E}0.14  & \cellcolor[HTML]{FCBB7A}0.72  & \cellcolor[HTML]{F8756D}0.11  \\
ID $\rightarrow$ EN       & \cellcolor[HTML]{F86D6B}0.04  & \cellcolor[HTML]{6DC17C}41.60 & \cellcolor[HTML]{FA9A74}0.43  & \cellcolor[HTML]{F86A6B}0.01  \\
EN $\rightarrow$ ZH       & \cellcolor[HTML]{F86D6B}0.04  & \cellcolor[HTML]{FBEA84}2.51  & \cellcolor[HTML]{8BCA7E}33.31 & \cellcolor[HTML]{F8696B}0.00  \\
ZH $\rightarrow$ EN       & \cellcolor[HTML]{FAEA84}2.58  & \cellcolor[HTML]{FAEA84}2.71  & \cellcolor[HTML]{F86F6C}0.06  & \cellcolor[HTML]{C7DB81}16.87 \\
None           & \cellcolor[HTML]{FAEA84}2.60   & \cellcolor[HTML]{FDD880}0.97  & \cellcolor[HTML]{FCC37C}0.79  & \cellcolor[HTML]{FDC87D}0.83 \\
\hline
\end{tabular}
  \caption{\label{tab:uni_result}
    Translation performance (BLEU scores) of adding parallel data at the end of training. Training with parallel data of all directions is synonymous with the \textsc{Parallel Last (all)} training setup, while using no parallel data (\textsc{None}) is synonymous with the \textsc{No Parallel} training setup. BLEU scores on the diagonal correspond to the \textsc{Parallel Last (uni)} training setup.
  }
\end{table}

\begin{table}[thb]
\setlength\tabcolsep{1.5pt}
\centering
\begin{tabular}{p{0.3\linewidth} | R{0.18\linewidth} R{0.23\linewidth} R{0.18\linewidth}}
\hline
{}  & \multicolumn{3}{c}{Common-Sense Reasoning} \\
\cline{2-4}
Parallel Data & English & Indonesian & Chinese \\
\hline
All directions      & 49.08 & {56.06}                         & {50.55}                     \\
EN $\rightarrow$ ID & 48.77 & {55.34} & {\color[HTML]{333333} 45.87}                                  \\
ID $\rightarrow$ EN & 48.73 & {55.87}             & {\color[HTML]{333333} 45.47}                                  \\
EN $\rightarrow$ ZH & 49.80 & {\color[HTML]{333333} 49.96}                                  & {50.49} \\
ZH $\rightarrow$ EN & 48.49 & {\color[HTML]{333333} 50.02}                                  & {50.45}         \\
\hline
\end{tabular}
  \caption{\label{tab:uni_cs}
    Macro-average scores on common-sense reasoning benchmarks for different translation directions of parallel training data. Training with parallel data of all directions is synonymous with the \textsc{Parallel Last (all)} training setup.
  }
\end{table}

\subsection{Quality of Translation Pairs}
Next, we investigate the effect of translation pair quality on LLMs' translation capability. Parallel data are sometimes collected automatically, leading to varying degrees of quality. In traditional machine translation systems, noisy translation pairs are often filtered out from the training data~\cite{Lowphansirikul2020ALE}. However, LLMs are typically trained on massive datasets that are inherently noisy. In this analysis, we examine whether filtering the parallel data could further enhance the multilingual performance of LLMs.  

\begin{table}[thb]
\setlength\tabcolsep{1.5pt}
\centering
\begin{tabular}{p{0.3\linewidth} | R{0.14\linewidth} R{0.14\linewidth} R{0.14\linewidth} R{0.14\linewidth} }
\hline
{}  & \multicolumn{4}{c}{Translation Evaluation} \\
\cline{2-5}
Parallel Data & \multicolumn{1}{L{0.14\linewidth}}{EN $\rightarrow$ ID} & \multicolumn{1}{L{0.14\linewidth}}{ID $\rightarrow$ EN} & \multicolumn{1}{L{0.14\linewidth}}{EN $\rightarrow$ ZH} & \multicolumn{1}{L{0.14\linewidth}}{ZH $\rightarrow$ EN} \\
\hline
\multicolumn{5}{c}{Zero-Shot} \\
\hline
\textsc{Parallel Last (all)} & 35.91 & 35.36 & 9.62  & 10.73 \\
\enspace with filtering & 36.20 & 35.17 & 24.20  & 10.60 \\
\hline
\multicolumn{5}{c}{Few-Shot (5 examples)} \\
\hline
\textsc{Parallel Last (all)} & 41.51 & 39.08 & 32.61 & 15.20  \\
\enspace with filtering & 40.50 & 37.71 & 32.28 & 14.83  \\
\hline
\end{tabular}
  \caption{\label{tab:filter_result}
    Translation performance (BLEU scores) of LLMs in relation to the quality of the parallel data.
  }
\end{table}

We assess the quality of the parallel data using CometKiwi-2022~\cite{rei-etal-2022-cometkiwi}, a state-of-the-art quality estimation model for machine translation. CometKiwi-2022 produces a score between 0 and 1, indicating the quality of each translation pair. Based on manual observation, we set a threshold of 0.42 for Chinese-English pairs and 0.58 for Indonesian-English pairs. Applying this filter reduces the number of parallel sentences from 33.9M to 25M for Chinese and from 54.1M to 15.6M for Indonesian. In this setting, we keep the amount of non-parallel data constant, resulting in the filtered model being trained on slightly less total data.  

We conduct this experiment using the \textsc{Parallel Last (all)} strategy. In zero-shot evaluation, we observe a significant improvement in English-Chinese translation performance and a slight improvement in English-Indonesian translation performance, but not in the other language pairs (Table~\ref{tab:filter_result}). In few-shot evaluation, the performance of the filtered model is slightly worse than that of the unfiltered model. This suggests that at this scale, LLMs are quite resilient to noisy data.  

\subsection{Catastrophic Forgetting}
\begin{figure}[htb]
\centering
\includegraphics[width=0.99\linewidth]
{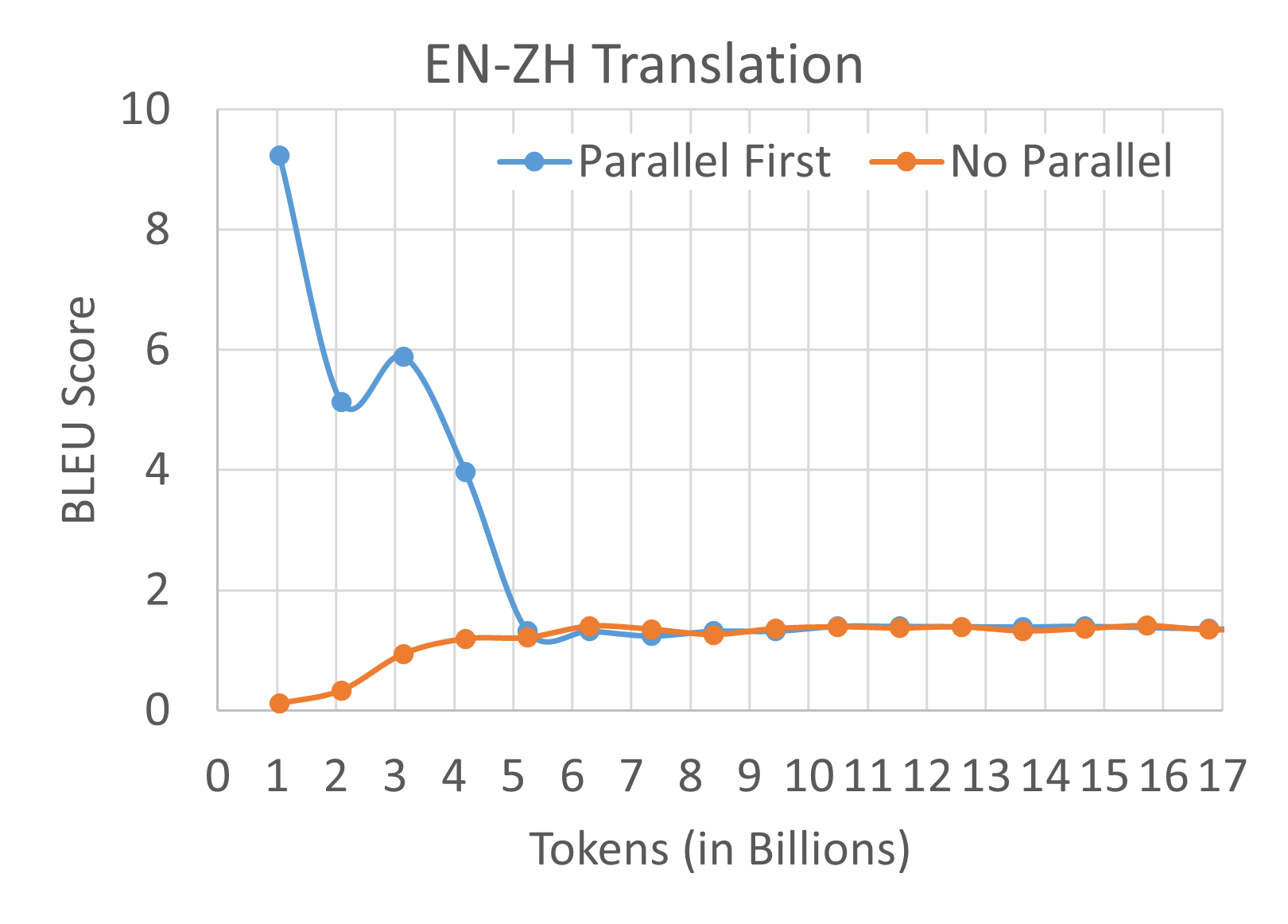}
\caption{Progression of the BLEU score on English-Chinese translation between \textsc{No Parallel} and \textsc{Parallel First} experimental settings.}
\label{fig:mt-front}
% \vspace*{-0.5\baselineskip}
\end{figure}

Adding parallel data at the beginning (\textsc{Parallel First}) improves LLMs' translation capability, but few-shot examples are needed to achieve considerable performance (Table~\ref{tab:result}). However, this performance is measured using the best checkpoint on the development set. When examining the performance at later checkpoints, we observe that the translation capability completely disappears once the parallel data are exhausted (Figure~\ref{fig:mt-front}). The only difference between the \textsc{Parallel First}, \textsc{Parallel Distributed}, and \textsc{Parallel Last} strategies is the position of the parallel data within the training set, yet placing it in the wrong position can completely negate its benefits. We attribute this to catastrophic forgetting~\cite{MCCLOSKEY1989109}. Similar patterns are observed in other translation directions.

\subsection{Impact of Incidental Bilingual Signals}
In this section, we investigate how incidental bilingual signals in the training data affect the LLMs' translation performance. \citet{briakou-etal-2023-searching} reported that incidental bilingual signals, especially parallel text, in predominantly English training data contribute to LLMs' translation abilities. However, they did not examine the relationship between the frequency of these bilingual signals and translation performance. In our analysis, we aim to extend their findings by exploring how the frequency of bilingual signals in our subset of SlimPajama influences the translation performance of models trained without parallel data.

\begin{figure}[thb]
\centering
\includegraphics[width=0.99\linewidth]
{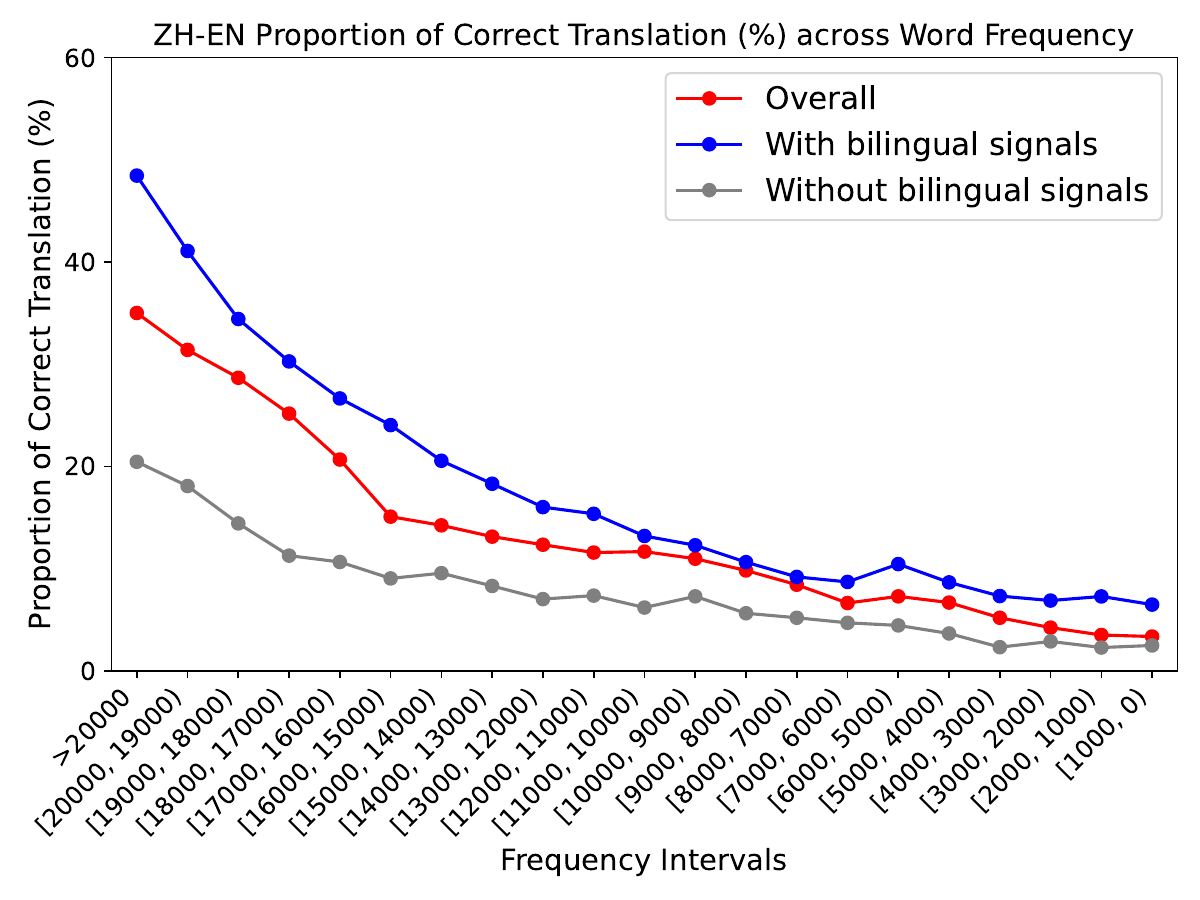}
\caption{Chinese-to-English translation performance across word frequency in the 167B training data.}
\label{fig:zh-en-frequency}
\end{figure}

\begin{figure}[thb]
\centering
\includegraphics[width=0.99\linewidth]
{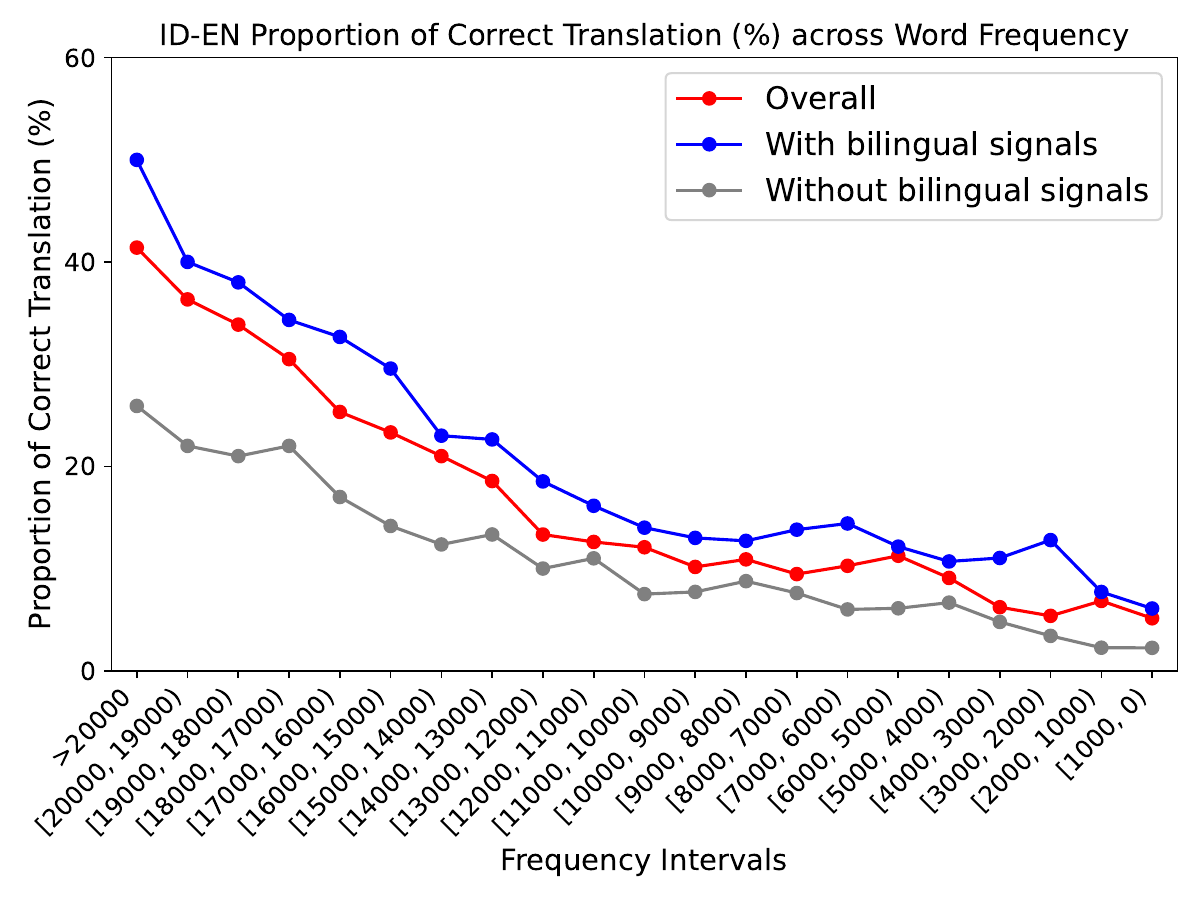}
\caption{Indonesian-to-English translation performance across word frequency in the 167B training data.}
\label{fig:id-en-frequency}
\end{figure}

We detect bilingual signals by first building a word translation dictionary by aligning words in the source and target sentences of the WMT-2022 test set (for Chinese-English) and the Flores-200 dev set (for Indonesian-English) using SimAlign~\cite{simalign}. Before processing with SimAlign, we perform word segmentation using jieba\footnote{https://github.com/fxsjy/jieba} for Chinese and NLTK~\cite{bird-loper-2004-nltk} for Indonesian and English. For each word in these test sets, we examine its context and frequency in the training data using Elasticsearch. We then categorize the words into two groups: words with bilingual signals and words without bilingual signals. A word is considered to have bilingual signals if it appears together with its translation in the same 2048-token context.

Next, we measure the translation accuracy of each word by checking whether its translation appears in the model's predictions. We then analyze the relationship between the frequency of Chinese or Indonesian words and their correct translation ratio, as shown in Figure~\ref{fig:zh-en-frequency} and Figure~\ref{fig:id-en-frequency}. In these figures, the x-axis represents the word frequency in the 167B tokens of training data. We group the words into 21 frequency intervals and calculate the average correct translation ratio for each interval.

The results indicate that words with bilingual signals achieve a higher correct translation ratio compared to words without bilingual signals. This demonstrates that the amount of bilingual signals in the training data significantly impacts the LLM's translation performance, which motivates our main experiment.

\subsection{Recommendations}
In this section, we briefly discuss our key findings for the future development of LLMs.
\begin{enumerate}
    \item \textbf{Leverage parallel data} \\
    When building multilingual large language models, parallel data should be fully utilized. We believe that reducing parallel data to merely monolingual data in other languages by scattering it throughout the training data—as done by BLOOM—is wasteful. The information contained in parallel data should be maximally leveraged by preserving its parallel format.
    \item \textbf{Second-stage training} \\
    In addition to maintaining the format of parallel data, the timing of its introduction to the model also plays a significant role. \citet{gururangan-etal-2020-dont} reported that second-phase pre-training on task-specific domains can improve a model’s performance on those tasks. From our systematic study, we found that training a model with parallel data after non-parallel data not only improves translation performance but also enhances multilingual common-sense reasoning.
    
    This strategy has been adopted by recent multilingual LLMs such as Tower \cite{alves2024tower} and Pangea \cite{yue2025pangea}, to some extent, in the form of instruction tuning with parallel data from machine translation tasks or by augmenting multimodal instruction tuning data with their translations. We expect to see broader utilization of parallel data in future multilingual LLMs.
    \item \textbf{Specialized translation systems} \\
    Fine-tuning a large language model using parallel data, even though it is pre-trained primarily on English data, offers a quick and effective method for developing high-quality machine translation systems. This approach has gained popularity recently \cite{xu2025xalma, Zeng_Meng_Yin_Zhou_2024}. Furthermore, we found that the model can achieve even higher performance by formatting the parallel data solely in the desired translation direction.
\end{enumerate}

\section{Conclusion}
In this work, we report a systematic study of the effect of including parallel data in the training data on large language models' multilingual capabilities, specifically focusing on translation and multilingual common-sense reasoning. Using English, Chinese, and Indonesian as case studies, we conduct controlled experiments to compare training large language models with mainly English data, with monolingual corpora of other languages, and with parallel data. We found that training LLMs with parallel data significantly enhances LLMs' multilingual capabilities.

Furthermore, we investigate how the location of parallel data affects the multilingual capabilities. We found that training the model with parallel data at the beginning of the training process is the least effective and leads to serious catastrophic forgetting. Conversely, training the model with parallel data at the end of the training process is the most effective in enhancing the model's multilingual capabilities. It significantly improves translation scores and also enhances the model's common-sense reasoning in other languages. Therefore, it is more effective to incorporate parallel data in second-stage training rather than randomly mixing them with non-parallel data.

Training the model in one translation direction can improve translation performance more than training in all translation directions simultaneously. However, this comes with a caveat: the model becomes totally unable to perform translation in other directions, including the opposite of what it was trained on.

Our source code, models, and data are publicly available to support further research into the multilingual capabilities of large language models. We hope that parallel data will be effectively leveraged in developing future open multilingual LLMs.

\section*{Limitations}
Due to resource constraints, we experiment with a language model containing 1.1B parameters, trained on 167B tokens, focusing on multilingual capabilities in English, Chinese, and Indonesian. Exploring additional languages and larger-scale models is left for future work.

Our experiments only use publicly available data. We believe it does not pose any direct societal risks.

\section*{Acknowledgments}

This research is supported by the National Research Foundation Singapore under its AI Singapore Programme (Award Number: AISG3-RP-2022-030).

\bibliography{custom}

\appendix

\section{Appendix}
\label{sec:appendix}

\begin{table*}[ht]
    \centering
    \begin{tabular}{cr|cr|cr}
        \hline\hline
        Language & Percentage & Language & Percentage & Language & Percentage \\ \hline\hline
        EN    &     82.35     &     PL     &    0.52      &  CS        &    0.18     \\ 
        FR    &    2.46     &     CA     &     0.51     &  TR        &    0.18      \\
        DE    &    2.13      &     CEB     &    0.34      &  DA        &    0.18      \\
        ES    &    1.74      &     FI     &    0.29      &   JA       &    0.18      \\
        IT    &      1.54    &     HU     &    0.23      &    FA      &    0.16      \\
        SV    &      0.92    &      AR    &    0.22      &    ALS      &   0.15       \\
        PT    &     0.86     &     ID     &    0.19      &    ZH      &   0.12       \\
        RU    &    0.74      &     EO     &     0.19     &     SR     &  0.12        \\
        NL    &     0.66     &     NO     &     0.18     &     HR     &   0.11       \\
        UK   &    0.60      &      KN    &      0.18    &     LA     &   0.11       \\ \hline\hline
    \end{tabular}
    \caption{Percentage (\%) of different languages (larger than 0.1\%) in our non-parallel training corpus.}
    \label{tab:lang_percentage}
\end{table*}

\subsection{Language Detection}

To measure the proportion of different languages in the training data, we use fastText\footnote{\url{https://fasttext.cc/docs/en/language-identification.html}}~\cite{joulin2016fasttext, joulin2017bag} to detect each word's language. For each word, fastText outputs a probability distribution over languages, and we assign its language based on the highest predicted probability. If a word is tokenized into multiple subwords by the tokenizer (e.g., BPE), we attribute the number of subwords to the original word’s detected language when computing language proportions. Table~\ref{tab:lang_percentage} presents the proportions of thirty languages in the 167B training data (each denoted by its ISO language code) with a proportion higher than $0.1\%$.

\subsection{Hyper-Parameters Setting}
\label{sec:hyper-param}
Our experiments use the TinyLlama 1.1B model, with architectural details provided in Table \ref{tab:arch-1b}. The hyper-parameters used for training are reported in Table \ref{tab:hyper-parameter}.
\begin{table}[thb]
\centering
\begin{adjustbox}{max width=0.4\textwidth}
\begin{tabular}{cc}
\hline
\hline
\textbf{Hyper-parameter} & \textbf{Value} \\
\hline
\hline
Number of Layers    & 22    \\
Embedding Dimension & 2048   \\
Intermediate Dimension & 5632    \\
Attention Heads & 32 \\
Query Groups & 4  \\
Context Window & 2048   \\
Vocabulary Size & 32000 \\
\hline
\hline
\end{tabular}
  \end{adjustbox}
  \caption{\label{tab:arch-1b}
    Architecture of our model.
  }
\end{table}

\begin{table}[htb]
\centering
\begin{adjustbox}{max width=0.4\textwidth}
\begin{tabular}{cc}
\hline
\hline
\textbf{Hyper-parameter} & \textbf{Value} \\
\hline
\hline
Number of GPUs    & 8    \\
Global Batch Size  & 512   \\
Micro Batch Size & 16 \\
Learning Rate & 4e-4   \\
Warmup Steps & 2000    \\
Weight Decay & 1e-1  \\
Optimizer & AdamW \\
~($\beta_1$, $\beta_2$) & (0.9, 0.95) \\
Gradient Clip & 1.0 \\
Minimal Learning Rate & 4e-5 \\
\hline
\hline
\end{tabular}
  \end{adjustbox}
  \caption{\label{tab:hyper-parameter}
    Hyper-parameters of our experiments.
  }
\end{table}

\begin{table}[htb]
    \centering
    \begin{tabular}{p{0.9\linewidth}}
    \hline
    Translate this from English to Indonesian \\
    English: The pilot was identified as Squadron Leader Dilokrit Pattavee. \\
    Indonesian: \\
    \hline
    \end{tabular}
    \caption{An example input prompt for English to Indonesian zero-shot translation evaluation.}
    \label{tab:zero_prompt_example}
\end{table}

\subsection{Common-Sense Reasoning Performance}
We provide the details of the models' performance on Chinese and Indonesian common-sense reasoning tasks in Table \ref{tab:zh-commonsense-result} and Table \ref{tab:id-commonsense-result} respectively.

\begin{table}[htb]
\centering
\begin{adjustbox}{max width=0.5\textwidth}
\begin{tabular}{lrrrrr}
\hline
\hline
\textbf{Model} & \textbf{XCOPA} & \textbf{XNLI} & \textbf{XStoryCloze} & \textbf{XWG} & \textbf{Avg} \\
\hline
\hline
No Parallel    & 52.20     & 33.33     & 48.64     & 53.37 & 46.88  \\
Multilingual   & 53.40     & 34.10     & 49.04     & 59.13 & 48.92 \\
Parallel Non-Adjacent & 51.00     & 33.57     & 49.37    & 59.13 & 48.27   \\
Parallel First        & 48.40     & 34.74     & 48.38    & 51.19 & 45.68  \\
Parallel Distributed  & 52.00     & 34.90     & 49.17    & 57.54 & 48.40   \\
Parallel Last (all)   & 54.40     & 33.73     & 52.15    & 61.90 & 50.55 \\
\hline
BLOOM          & 59.40     & 36.67     & 58.04      & 69.05 & 55.79 \\
\hline
\hline
\end{tabular}
  \end{adjustbox}
  \caption{\label{tab:zh-commonsense-result}
    Chinese common-sense reasoning performance (accuracy) in each experimental setting, measured based on the model's accuracy on XCOPA, XNLI, XStoryCloze, and XWinograd.
  }
\end{table}

\begin{table}[htb]
\centering
\begin{adjustbox}{max width=0.4\textwidth}
\begin{tabular}{lrrr}
\hline
\hline
\textbf{Model} & \textbf{XCOPA} & \textbf{XStoryCloze} & \textbf{Avg} \\
\hline
\hline
No Parallel    & 50.40    & 49.44 & 49.92    \\
Multilingual   & 52.60    & 50.36 & 51.48    \\
Parallel Non-Adjacent & 52.40    & 50.43 & 51.41 \\
Parallel First        & 52.20    & 49.90 & 51.05 \\
Parallel Distributed  & 55.20    & 52.88 & 54.04   \\
Parallel Last (all)   & 57.40    & 54.73 & 56.06  \\
\hline
BLOOM          & 64.60    & 57.78 & 61.19     \\
\hline
\hline
\end{tabular}
  \end{adjustbox}
  \caption{\label{tab:id-commonsense-result}
    Indonesian common-sense reasoning performance (accuracy) in each experimental setting, measured based on the model's accuracy on XCOPA and XStoryCloze.
  }
\end{table}

\end{document}